\documentclass[letterpaper]{article} 
\usepackage{aaai2026}  
\usepackage{times}  
\usepackage{helvet}  
\usepackage{courier}  
\usepackage[hyphens]{url}  
\usepackage{graphicx} 
\usepackage{natbib}  
\usepackage{caption} 
\usepackage{algorithm}
\usepackage{algorithmic}
\usepackage{subcaption}
\usepackage{amsmath}
\usepackage{multirow}
\usepackage{booktabs}
\newtheorem{assumption}{Assumption}
\usepackage{array}
\usepackage{newfloat}
\usepackage{listings}
\DeclareCaptionStyle{ruled}{labelfont=normalfont,labelsep=colon,strut=off} 
\floatstyle{ruled}
\newfloat{listing}{tb}{lst}{}
\floatname{listing}{Listing}
\title{SACO: Sequence-Aware Constrained Optimization Framework for Coupon Distribution in E-commerce}
\author{
    Li Kong\textsuperscript{\rm 1}\equalcontrib,
    Bingzhe Wang\textsuperscript{\rm 1}\equalcontrib,
    Zhou Chen\textsuperscript{\rm 2},
    Suhan Hu\textsuperscript{\rm 1},
    Yuchao Ma\textsuperscript{\rm 1},
    Qi Qi\textsuperscript{\rm 1,4,5}\thanks{Corresponding Author.},
    Suoyuan Song\textsuperscript{\rm 3},
    Bicheng Jin\textsuperscript{\rm 3}
}
\affiliations{
    \textsuperscript{\rm 1}Gaoling School of Artificial Intelligence, Renmin University of China, China\\
    \textsuperscript{\rm 2}School of Economics and Management, Southeast University, China \\
    \textsuperscript{\rm 3}ByteDance, China \\
    \textsuperscript{\rm 4}Beijing Key Laboratory of Research on Large Models and Intelligent Governance \\
    \textsuperscript{\rm 5}Engineering Research Center of Next-Generation Intelligent Search and
Recommendation, MOE \\
    beverly\_kong@163.com, \{wbz2022, 2023202249, yc.ma, qi.qi\}@ruc.edu.cn,\\ chenzhou@seu.edu.cn, \{songsuoyuan, jinbicheng\}@bytedance.com
}

\usepackage{bibentry}

\begin{document}

\maketitle

\begin{abstract}
Coupon distribution is a critical marketing strategy used by online platforms to boost revenue and enhance user engagement. Regrettably, existing coupon distribution strategies fall far short of effectively leveraging the complex sequential interactions between platforms and users. This critical oversight, despite the abundance of e-commerce log data, has precipitated a performance plateau. In this paper, we focus on the scene that the platforms make sequential coupon distribution decision multiple times for various users, with each user interacting with the platform repeatedly. Based on this scenario, we propose a novel marketing framework, named \textbf{S}equence-\textbf{A}ware \textbf{C}onstrained \textbf{O}ptimization (SACO) framework, to directly devise coupon distribution policy for long-term revenue boosting. SACO framework enables optimized online decision-making in a variety of real-world marketing scenarios. It achieves this by seamlessly integrating three key characteristics, general scenarios, sequential modeling with more comprehensive historical data, and efficient iterative updates within a unified framework. Furthermore, empirical results on real-world industrial dataset, alongside public and synthetic datasets demonstrate the superiority of our framework.
\end{abstract}


\section{Introduction}

\begin{figure}[t]
    \centering
    \includegraphics[width=0.95\linewidth]{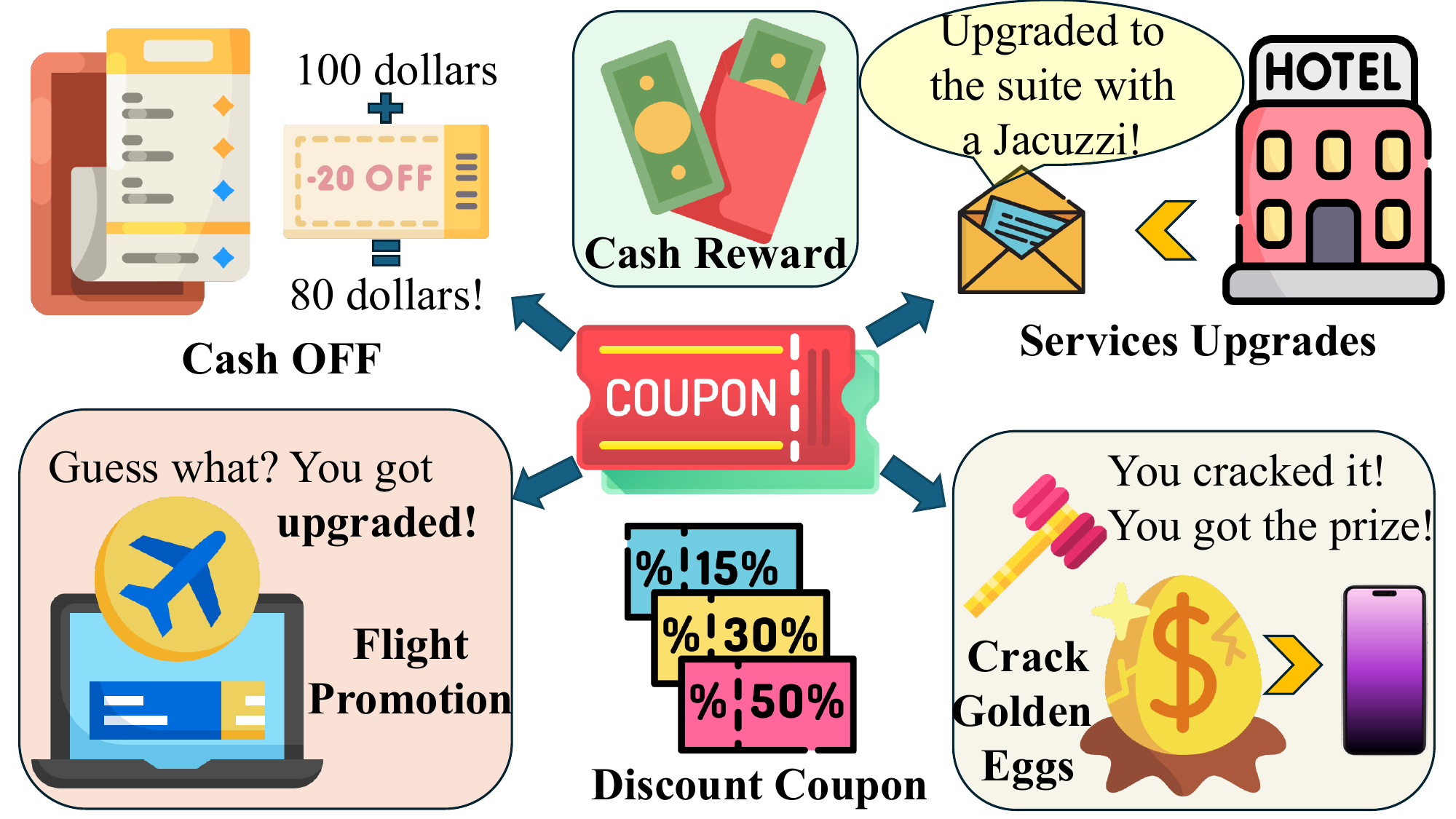}
    \caption{Complex and Meaningful Coupon Distribution.}
    \label{fig:coupon}
\end{figure}

In the highly competitive online ecosystem, coupon distribution has emerged as a central marketing strategy for online platforms looking to amplify user engagement and optimize revenue streams. Coupons take many forms (as shown in Figure \ref{fig:coupon}), from traditional discounts to service upgrades and cash rewards, all designed to incentivize consumption across diverse sectors. For instance, Booking's promotional activities can enhance user satisfaction \cite{goldenberg2020free, albert2022commerce}; Airbnb employs dynamic pricing \cite{ye2018customized}; Uber's promotions encourage users to adopt new products \cite{du2019improve}; Taobao's discounts increase user activity \cite{zhao2019unified, zhang2021bcorle}; Kuaishou offers cash rewards to stimulate user retention \cite{ai2022lbcf}; and Meituan utilizes spend\&save discounts to incentivize purchases \cite{zhou2023direct, huang2024entire, zhou2024decision}. 

However, these marketing campaigns come with significant costs. Typically, dedicated budgets are set to ensure campaign sustainability. Effective coupon distribution within these finite budgets is therefore a critical challenge, frequently formalized as a constrained optimization problem. Unlike traditional resource allocation, coupon distribution under budget constraints faces two significant challenges. First, user responses are variable and the counterfactual outcomes are fundamentally unobservable. Second, user responses are sequential, which means past coupon decisions affect users' subsequent responses, making counterfactual prediction challenging. As a result, it is impossible to obtain a direct solution.

\begin{figure}[t]
    \centering
    \includegraphics[width=1.0\linewidth]{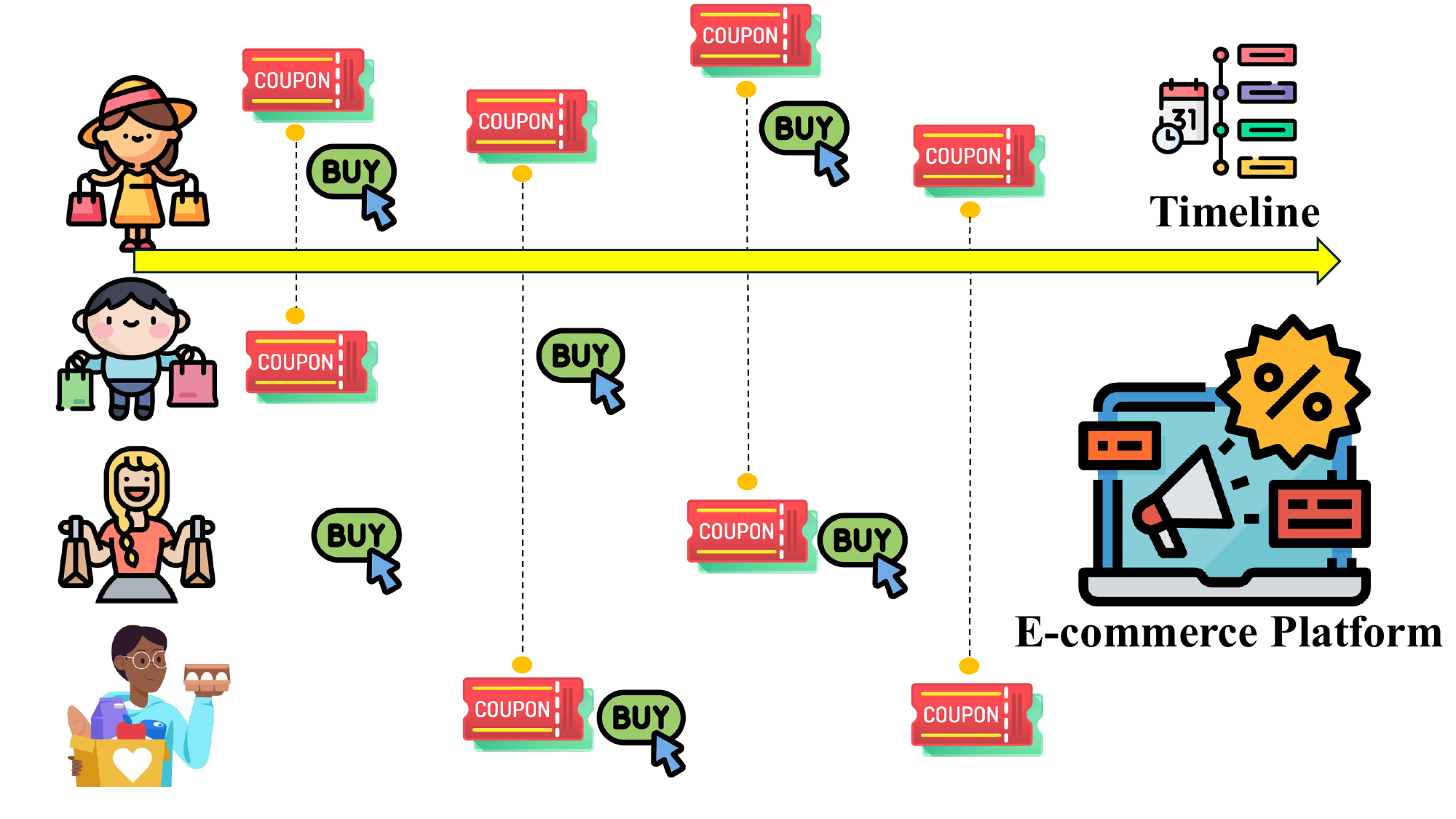}
    \caption{Single-Round Scenarios in Existing Research.} 
    \label{fig:scenario}
\end{figure}

While prior research has explored counterfactual policy optimization from logged bandit feedback \cite{swaminathan2015counterfactual, swaminathan2015batch, joachims2018deep}, the bandit approach relies on online exploration, rendering it unsuitable for coupon distribution in e-commerce scenarios where decisions, once made, are difficult to reverse. In contrast, more recent studies acknowledge the importance of pre-training models with logged data before deployment, representing a significant improvement.
However, they typically focus on scenarios where the platform makes a single-round coupon allocation decision for each user interaction according to the current state.
The prevailing two-stage approach dominates coupon allocation research \cite{hansotia2002direct, zhao2019unified, goldenberg2020free}, comprising: (1) machine learning (ML) prediction of user response to different coupons, and (2) optimization research (OR) for revenue maximization. However, decoupling the process into two stages leads to error accumulation, and the two-stage goals cannot be fully aligned.
Recent end-to-end frameworks, including decision-focused learning (DFL) \cite{zhou2023direct, zhou2024decision} and reinforcement learning (RL) \cite{xiao2019model, zhang2021bcorle}, attempt to address this limitation. However, these approaches face inherent performance limitations when constrained to this single-round interaction paradigm, as illustrated in Figure \ref{fig:scenario}. The top part visualizes the RL methods modeling the interactions with single user as Markov processes. In contrast, the bottom part shows Two-stage and DFL methods making decisions based on current state alone.
In these settings, the potential for leveraging the richer sequential dependencies in user behavior is significantly curtailed, leading to a performance plateau.

Unlike these single-round paradigms, our work, while inheriting the benefits of an end-to-end framework, embraces the sequential nature of user engagement within e-commerce platforms. We propose an efficient framework named \textbf{S}equence-\textbf{A}ware \textbf{C}onstrained \textbf{O}ptimization (SACO) framework to directly devise coupon distribution policy for long-term revenue boosting. Critically, while demonstrated here for coupon distribution, SACO's general framework empowers exploration of constrained optimization scenarios across many applications. Our contributions are threefold:

\textbf{Generalized and Practical Problem Setting.} We address more complex scenarios where: (1) multiple users arrive simultaneously per round (e.g., group-buying scenarios), requiring parallel coupon allocation; and (2) users may revisit the platform across multiple rounds, creating sequential dependencies.

\textbf{Efficient Framework.} SACO framework offers two key advantages: (1) by explicitly modeling the series of user interactions and utilizing historical e-commerce data, SACO captures valuable temporal dynamics overlooked by single-round approaches; and (2) by reformulating sequential decision-making as an auto-regressive prediction task, we leverage causal-Transformer structure to effectively capturing multi-round causal relationships. 
Meanwhile, this structure reduces the framework's reliance on the statistical premise of unbiased data, which serves as the training set foundation for two-stage methods and DFL based on uplift models. This allows for efficient, timely, and cost-effective iterative updates, contributing to sustained optimal performance in dynamic online environments.

\textbf{Extensive Experiments on Public, Synthetic and Industrial Datasets.} 
We validate our approach through comprehensive experiments on diverse datasets.
Comparisons against state-of-the-art baselines demonstrate an average revenue improvement of \textbf{3.60\%}. Supporting experiments include ablation studies and inference speed analysis.

\section{Further Related Work}

Coupon distribution requires estimating user responses to different coupon strategies, a task known as counterfactual evaluation. A key metric is the conditional average treatment effect (CATE), which quantifies the impact of a specific coupon on outcomes like conversion rates, considering individual user characteristics \cite{devriendt2018literature}. Uplift modeling, a field dedicated to CATE estimation, has produced various machine learning techniques, often trained on randomized controlled trial (RCT) data \cite{athey2015machine}. The trained models then predict individual-level causal effects for arbitrary interventions. Notable CATE estimation techniques include meta-learners \cite{hansotia2002direct, athey2015machine, kunzel2019metalearners, nie2021quasi}, ranking-based methods \cite{kuusisto2014support, betlei2021uplift}, representation learning \cite{johansson2016learning, shi2019adapting, yao2018representation}, and causal forests \cite{ai2022lbcf, athey2019generalized, wager2018estimation, zhao2017uplift}. However, existing CATE formulations primarily address single-treatment effects. In contrast, our problem requires modeling \textit{temporal treatment dependencies}, where user responses depend not only on the current coupon but also on historical coupon exposures. This introduces complex sequential counterfactual estimation challenges that traditional uplift models cannot address.

To tackle this challenge, we leverage the Transformer architecture \cite{vaswani2017attention}, which has demonstrated remarkable capabilities in sequential modeling across diverse domains including natural language processing \cite{keskar2019ctrl, dai2019transformer}, computer vision \cite{parvaiz2023vision}, and reinforcement learning (RL) \cite{chen2021decision, zheng2022online, wu2023elastic}. While prior RL applications focus on unconstrained optimization of cumulative rewards, we adapt the Decision Transformer (DT) framework \cite{chen2021decision} for constrained optimization problems. Our modification enables handling of budget constraints while preserving DT's ability to model sequential dependencies in decision-making.

\section{Coupon Distribution Problem}

\subsection{Model on Coupon Distribution}
\begin{figure*}[t]
    \centering
    \includegraphics[width=0.87\linewidth]{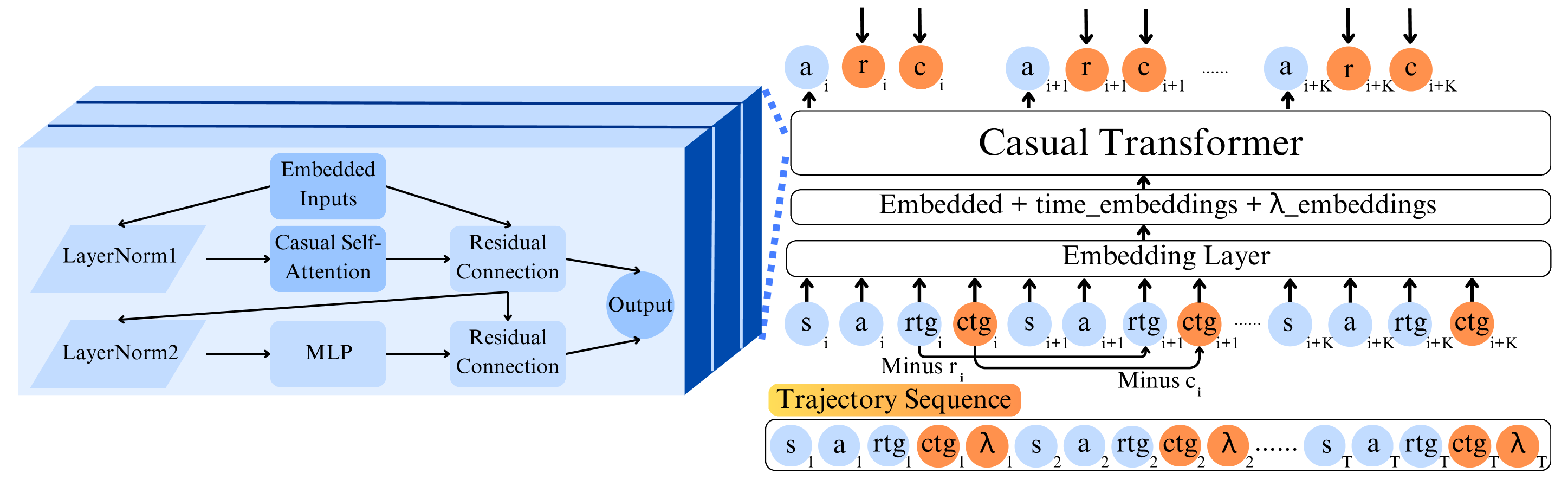}
    \caption{Constrained-Optimization Decision Transformer}
    \label{fig:model}
\end{figure*}

We formulate the coupon distribution problem as a budget allocation problem with multiple treatments, which was also introduced in many existing works. Different from them, we consider a comprehensive scenario with sequential coupon distribution decisions for diverse users who continuously interact with the e-commerce platform.

An e-commerce platform has $N$ users. Given the requirement that all users register on the platform before purchasing, $N$ is a known parameter. Over a finite time horizon, users interact with the platform for up to $T$ times, referred to as $T$ rounds. Upon receiving a user interaction request, the platform must make a real-time decision regarding coupon issuance. Critically, once a coupon is granted, it cannot be revoked.
Over the $T$ rounds, the platform has a marketing budget $B$, which is the total amount it is willing to spend on issuing coupons to all users. The platform offers $K$ types of coupons, such as $\{2,5,10\}$ dollar coupons for $K=3$, 
and so forth. Let $[N], [T], [K]$ denote the sets of users, rounds, and coupons, respectively. 

We define $x_{itk}$ as the platform's coupon issuance decision, where $x_{itk} = 1$ indicates that in round $t$, the platform issues coupon $k$ to user $i$, with $i \in [N], t \in [T], k \in [K]$. Following coupon issuance, a user generates consumption $r_{itk}$, while the platform incurs a corresponding cost of $c_{itk}$ as a subsidy.
The platform's objective is to maximize revenue from marketing within the limited budget $B$. Formally, we perform a linear relaxation on $x_{itk}$ and express the platform's online marketing problem as the following online optimization problem:
\begin{align}
\max \quad & \sum_{i \in [N], t \in [T], k \in [K]}{r_{itk} x_{itk}} \\
\text{s.t.} \quad & \sum_{i \in [N], t \in [T], k \in [K]}{c_{itk} x_{itk}} \le B, \\
& 0 \le x_{itk} \le 1, \forall i \in [N], t \in [T], k \in [K].
\end{align}

In previous research, the platform receives a request from only one user per round. In this paper, we generalize this constraint by allowing the platform to receive requests from multiple users per round. For instance, in group-buying scenarios, the platform needs to make coupon issuance decisions for multiple users simultaneously.
Furthermore, by leveraging Transformer architectures to process multi-dimensional action space, we allow the platform to issue a bundle of coupons (e.g. the combination of 5-dollar coupon and 10\% discount) to each user per time, which aligns more closely with real-world complex scenarios. 

\subsection{Dual Problem Formulation}
Let $\boldsymbol{x}$ represent the coupon distribution strategy, $R(\boldsymbol{x})$ denote the total revenue obtained by the platform under this strategy, and $C(\boldsymbol{x})$ represent the total amount of coupons subsidized by the platform under this strategy. To obtain the optimal distribution strategy within the budget constraint, we first introduce $\lambda$ as the dual variable for the constraint $\sum_{i \in [N], t \in [T], k \in [K]}{c_{itk} x_{itk}} \le B$. Next, we define the Lagrangian function $L(\boldsymbol{x}, \lambda)$ and the objective function of the dual problem is the maximum value of the Lagrangian function with respect to $\boldsymbol{x}$, i.e.,
\begin{align}
& \min_{\lambda \ge 0}{\max_{0 \le x_{itk} \le 1}{ L(\boldsymbol{x}, \lambda)}} \\
= & \min_{\lambda \ge 0}{\max_{0 \le x_{itk} \le 1}{ \left( R(\boldsymbol{x}) - \lambda \left( C(\boldsymbol{x}) - B \right) \right)}}
\end{align}

Problem (1) is formulated as a linear program. Therefore, strong duality holds, implying that the we can find the optimal primal solution by solving the dual problem. Given optimal dual variable $\lambda^{*}$ and known parameter $B$, the optimization objective simplifies to 
\begin{align}
\max_{0 \le x_{itk} \le 1}{ \left( R(\boldsymbol{x}) - \lambda^{*} C(\boldsymbol{x}) \right)}.
\end{align}

Assuming all responses are known, the problem simplifies to a classical multiple choice knapsack problem, yielding an approximation ratio near 1\cite{zhou2023direct, zhou2024decision}.
However, the platform can only observe the actual revenue or cost after a specific coupon decision and the counterfactual data is inherently unattainable. 
Furthermore, user responses are sequential. Previous coupon distribution decisions influence subsequent users' coupon responses, making counterfactual prediction challenging.
Therefore, the online linear programming problem cannot be directly solved. 

\section{Our Framework}
To address the above challenge, we propose our framework named \textbf{S}equence-\textbf{A}ware \textbf{C}onstrained \textbf{O}ptimization (SACO) framework, a direct policy learning approach involving training a model to determine the optimal sequential policy  $\boldsymbol{x_i}$ for each user $i$ given $\lambda$, then leveraging a $\lambda$ optimization algorithm and model-parallel inference to derive the optimal overall strategy  $\boldsymbol{x}$. 

\subsection{Trajectory Representation}
In the e-commerce platform, learning policies through direct interaction with the online environment is infeasible. However, by organizing historical online logs into structured trajectories, we unlock a feasible strategy for effective model training. We model the sequential decision-making problem for single-user and propose the trajectory representation as the training dateset format. 

The sequential decision-making problem is represented by the tuple $<\mathcal{S}, \mathcal{A}, \mathcal{P}, \mathcal{R}, \mathcal{C}, \gamma>$. Specifically,

\begin{itemize}
\item $\mathcal{S}$: The state space, which describes the user context, including user features and other relevant information. $s_{t} \in \mathcal{S}$ represents the user context for all users in round $t$, where $\boldsymbol{s}_{t} = (s_{1t}, \ldots, s_{it}, \ldots, s_{Nt})$.
\item $\mathcal{A}$: The coupon distribution action space, which encompasses the choices of treatments for users. $\boldsymbol{a}_{t} \in \mathcal{A}$ represents the coupon distribution strategy in round $t$, where $\boldsymbol{a}_{t} = (x_{1t1}, \ldots, x_{Nt1}, \ldots, x_{itk}, \ldots, x_{NtK})$.
\item $\mathcal{P}$: The state transition function, $\mathcal{P}: \mathcal{S} \times \mathcal{A} \rightarrow \mathcal{S}$. The transition dynamics $P(s_{t+1} | s_{t}, a_{t})$ describe the dynamic transition from the current state $s_{t}$ to the next state $s_{t+1}$ after taking an action $a_{t}$.
\item $\mathcal{R}$: The reward space, $\mathcal{R}: \mathcal{S} \times \mathcal{A} \rightarrow \mathcal{R} \in {R}$, which describes the platform revenue generated by coupon distribution. $r_{t} \in \mathcal{R}$ represents the platform revenue in round $t$, where $r_{t} = \sum_{i \in [N]}{r_{it}}$ and $r_{it} = \sum_{k \in [K]}{r_{itk}}$. 
\item $\mathcal{C}$: The cost space, $\mathcal{C}: \mathcal{S} \times \mathcal{A} \rightarrow \mathcal{C} \in {R}$, which describes the marketing budget expenditure incurred by coupon distribution actions. $c_{t} \in \mathcal{C}$ represents the marketing budget expenditure in round $t$, where $c_{t} = \sum_{i \in [N]}{c_{it}}$ and $c_{it} = \sum_{k \in [K]}{c_{itk}}$.
\item $\gamma$: The discount factor for future rewards, indicating the importance of future rewards relative to the present. In this paper, we set $\gamma = 1$.
\end{itemize}

Given a dataset in the format (user-id, user-features as state, treatment as action, cost, reward, time) and additional $\lambda$ randomized in the interval $(0,1)$, we aggregate the dataset by user-id and then sort the data with each user's group in chronological order. We define $RTG_{i_t} = \sum_{t'=t}^T \gamma r_{it}$ \cite{chen2021decision} and $CTG_{i_t} = \sum_{t'=t}^Tc_{it}$. The training dataset is pre-processed into trajectory data of size $N$ :
\begin{align*}
\tau_{i\in [N]} = & [(s_{i_0},a_{i_0},CTG_{i_0}, RTG_{i_0},\lambda_i), \ldots, \\
& [(s_{i_t},a_{i_t},CTG_{i_t}, RTG_{i_t},\lambda_i), \ldots,\\
& (s_{i_T},a_{i_T},CTG_{i_T},RTG_{i_T},\lambda_i) ]
\end{align*}

As mentioned above, the dual variable $\lambda$ reflects the constraint on the budget $B$. To improve the model's ability to generalize across diverse budget levels, we randomize $\lambda$ ten times for each trajectory sample. This expands our training dataset tenfold, promoting more robust learning.

\begin{algorithm}[tb]
\caption{Model Training}
\label{alg:decision-transformer} 
\textbf{Input}:State dimension $d_s$, action dimension $d_a$ \\
    Maximum episode length $T$, sequence length $K$ \\
\textbf{Parameter}: Learning rates $\eta$, weight decay $\beta$\\
\textbf{Output}: Trained model $\theta$
\begin{algorithmic}[1] 
\STATE \textbf{Initialize:}$\theta$, $\text{AdamW}(\theta, \eta, \beta)$\\
\FOR{epoch $= 1$ \textbf{to} $N_{\text{epochs}}$}
    \STATE Sample batch $\mathcal{B} = \{(s_t, a_t, \text{rtg}_t, \text{ctg}_t, t, \lambda)\}_{t=j}^{j+K}$\\
    \STATE Normalize states: $s_t \gets (s_t - \mu_s) / \sigma_s$\\
    \STATE \textbf{Compute embeddings:}
    \STATE $e_s \gets \text{Embed}(s_t) + \text{Embed}(t) + \text{Embed}(\lambda_t)$
    \STATE $e_a \gets \text{Embed}(a_t) + \text{Embed}(t) + \text{Embed}(\lambda_t)$
    \STATE $e_{\widehat{RTG}} \gets \text{Embed}(\widehat{RTG}_t) + \text{Embed}(t) + \text{Embed}(\lambda_t)$
    \STATE $e_{\widehat{CTG}} \gets \text{Embed}(\widehat{CTG}_t) + \text{Embed}(t) + \text{Embed}(\lambda_t)$
    \STATE Stack embeddings: $x \gets \text{Stack}(e_s, e_a,e_{\widehat{RTG}}, e_{\widehat{CTG}})$
    \STATE Apply layer norm: $x \gets \text{LayerNorm}(x)$
    \FOR{\textbf{each transformer block} $\text{Block}_i$}
        \STATE $x \gets \text{Block}_i(x)$
    \ENDFOR
    \STATE \textbf{Predict: $\hat{a}_t$}
    \STATE \textbf{Compute task losses: $\mathcal{L}_{a}$}
    \STATE Backpropagate $\mathcal{L}_{a}$ and update $\theta$ using optimizers
\ENDFOR
\STATE \textbf{return} Trained model $\theta$
\end{algorithmic}
\end{algorithm}

\subsection{Constrained-Optimization Decision Transformer}
Having established the training dataset, we adapt the Decision Transformer model and align it for direct policy learning, as summarized in Figure \ref{fig:model} and Algorithm \ref{alg:decision-transformer}. 

Our model leverages two key advantages of the causal transformer structure, the ability to handle sequential data effectively via its transformer architecture and the capability to evaluate counterfactual outcomes through its causal structure\cite{nichani2024transformers}. 

To integrate constraint information, we augment the input space with the parameters $ctg$, representing constraint types. Furthermore, we enhance the original time embedding $Embed(t)$ by incorporating the dual variable embedding $Embed(\lambda)$, as shown in line 6-9 of Algorithm \ref{alg:decision-transformer}. This enhancement effectively encoded constraint information alongside temporal signals for each parameter. Subsequently, these augmented inputs are processed through the Causal Transformer blocks, with the internal details visualized in the left panel of Figure \ref{fig:model}.

As shown in line 16 of Algorithm \ref{alg:decision-transformer}, we employed a cross-entropy loss calculated on the action predictions:
\begin{align}
\mathcal{L}_{a}(\theta) = -\frac{1}{NT} \sum_{i=1}^{N}\sum_{t=1}^{T} \sum_{j=1}^{K} y_{itj} \log(p_{\theta}(a_{it} = j' | s_{it})) 
\end{align}
only when $j=j'$, then $y_{itj}=1$, meaning that the $j'$ type of coupon is distributed to the user in the dataset and the response is recorded and observable.
This design choice enables the model to learn optimal policies tailored to specific optimization objectives and constraints.

Following training, we obtained the model capable of making optimal sequential decisions for individual users, conditioned on the dual variables. 

\subsection{Overall Strategy}
\begin{algorithm}[tb]
\caption{Overall Algorithm of SACO} 
\label{alg:optimize_lambda}
\textbf{Input}: $\text{N}$, user states $\boldsymbol{s}$, budget $B$\\
\textbf{Parameter}: $\text{max\_iterations}$, $\text{noise}$, \text{learning\_rate} $lr$\\
\textbf{Output}: $\lambda^*$, $R^*$
\begin{algorithmic}[1] 
\STATE \textbf{Initialize:}$\lambda \leftarrow \text{Uniform}(0, 1)$
\STATE \textbf{Define Function OBJECTIVE($\lambda$):}
\STATE \ \ \ \ Simulate all users with $\lambda$ through trained model
\STATE \ \ \ \ Compute total cost $C$ and total revenue $R$
\STATE \ \ \ \ $\text{penalty} \gets \lambda \cdot \max(C(\boldsymbol{x}) - B, 0)$
\STATE \ \ \ \  \textbf{return} $\text{penalty} - R(\boldsymbol{x})$
\WHILE{$C>B$ or $C<B-\epsilon$}
    \STATE Reset the offline simulation environment
    \STATE $\text{init} \leftarrow \lambda + \text{Uniform}(-\text{noise}, \text{noise})$
    \STATE $\text{init} \leftarrow \text{Clip}(\text{init}, 0, 1)$ 

    \STATE \textbf{Optimize $\lambda$ using SLSQP:}
    \STATE $\text{result} \leftarrow \text{minimize}(\text{OBJECTIVE}, \text{init}, \text{SLSQP}, \text{bounds})$
    \STATE $\lambda \leftarrow \text{result}.x[0]$ \COMMENT{Update $\lambda$}

    \STATE $\text{R}, \text{C} \leftarrow (\text{user\_states}, \lambda)$
    \IF{$\text{R} > \text{R*}$ \textbf{and} $\text{C} \leq \text{B}$}
        \STATE $\text{R*} \leftarrow \text{R}$, $\text{$\lambda^*$} \leftarrow \lambda$
    \ENDIF
    \STATE $\lambda \leftarrow \lambda +$lr$ \cdot (C-B)$ \COMMENT{Dynamic adjustment}
\ENDWHILE
\STATE \textbf{return} $\lambda^*$, $R^*$
\end{algorithmic}
\end{algorithm}

Based on the trained model to determine the optimal sequential policy  $\boldsymbol{x_i}$ for each user $i$ given $\lambda$, we combine it with the $\lambda$ optimization algorithm, obtaining the SACO framework to derive the optimal overall strategy $\boldsymbol{x}$, as summarized in Algorithm \ref{alg:optimize_lambda}.

We adopt the SLSQP algorithm to optimize Lagrangian multiplier $\lambda$, as it is versatile and robust. SLSQP eliminates the need for fine-tuning while delivering reliable solutions for practical engineering applications. 

We set the objective of SLSQP algorithm as 
\begin{equation}
    \lambda \cdot (C(\boldsymbol{x}) - B) - R(\boldsymbol{x}).
\end{equation}

Considering the limitation that the SLSQP algorithm may fall into the local optimum, we will perform multiple rounds of iterative optimization in our framework. In each new round of iterative optimization process, we will add some noise disturbance and penalty on $\lambda$ to appropriately guide the update direction, shown in line 9 and 18 of Algorithm \ref{alg:optimize_lambda}. 

\begin{assumption}
When $C(\boldsymbol{x})>C(\boldsymbol{x'})$, then $R(\boldsymbol{x})>R(\boldsymbol{x'})$.
\end{assumption}

We operate under the assumption of a direct correlation between reward magnitude and associated cost, an observation validated by typical marketing and advertising practices. Hence, we define the convergence criterion for iteration as $B-\epsilon <C<B$.

Finally, we obtain the overall framework to directly devise coupon distribution policy for long-term revenue boosting.

Unlike prior single-round approaches, SACO, as a sequential modeling framework, unlocks the potential to extract complex user-platform interaction patterns from rich datasets.
Critically, while demonstrated here for coupon distribution, SACO's general framework empowers exploration of constrained optimization scenarios across a spectrum of applications. 
Furthermore, its significant scalability allows for straightforward integration of external components.

\section{Experiment}
\begin{figure*}[ht!]
    \centering
    \begin{subfigure}[b]{0.246\textwidth}
        \includegraphics[width=\linewidth]{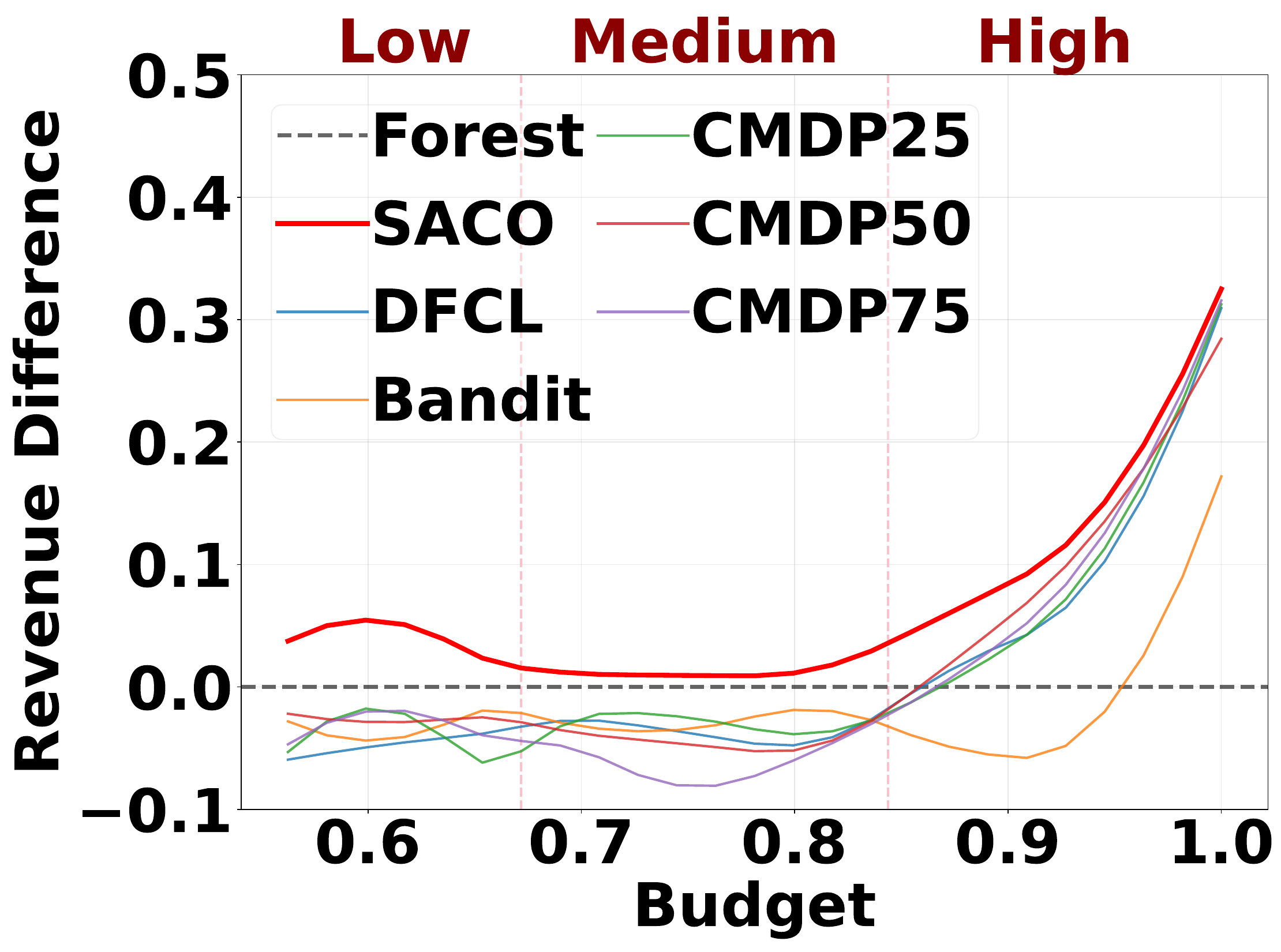}
        \caption{CU}
    \end{subfigure}
    \hfill
    \begin{subfigure}[b]{0.246\textwidth}
        \includegraphics[width=\linewidth]{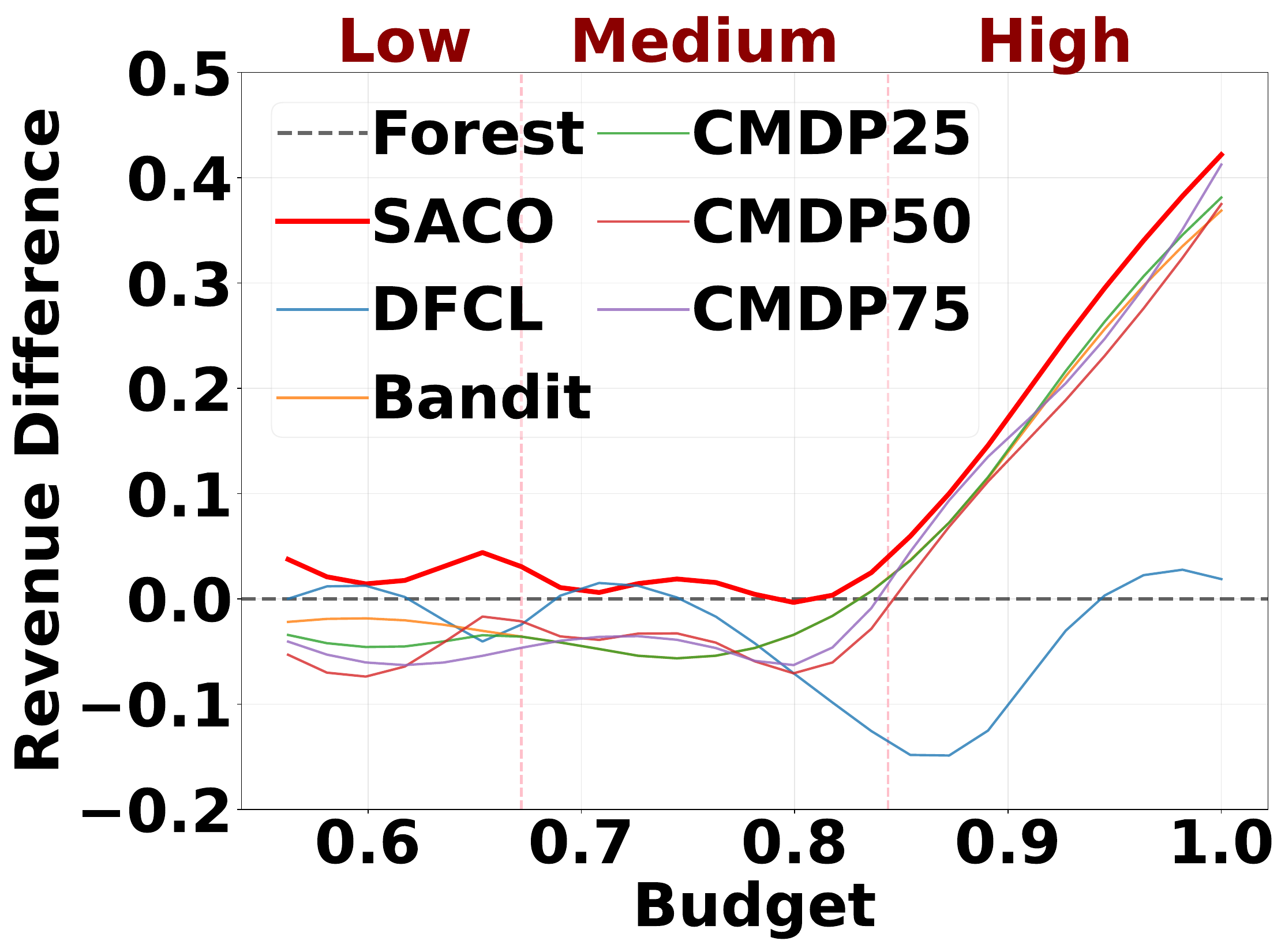}
        \caption{MT}
    \end{subfigure}
    \hfill
    \begin{subfigure}[b]{0.246\textwidth}
        \includegraphics[width=\linewidth]{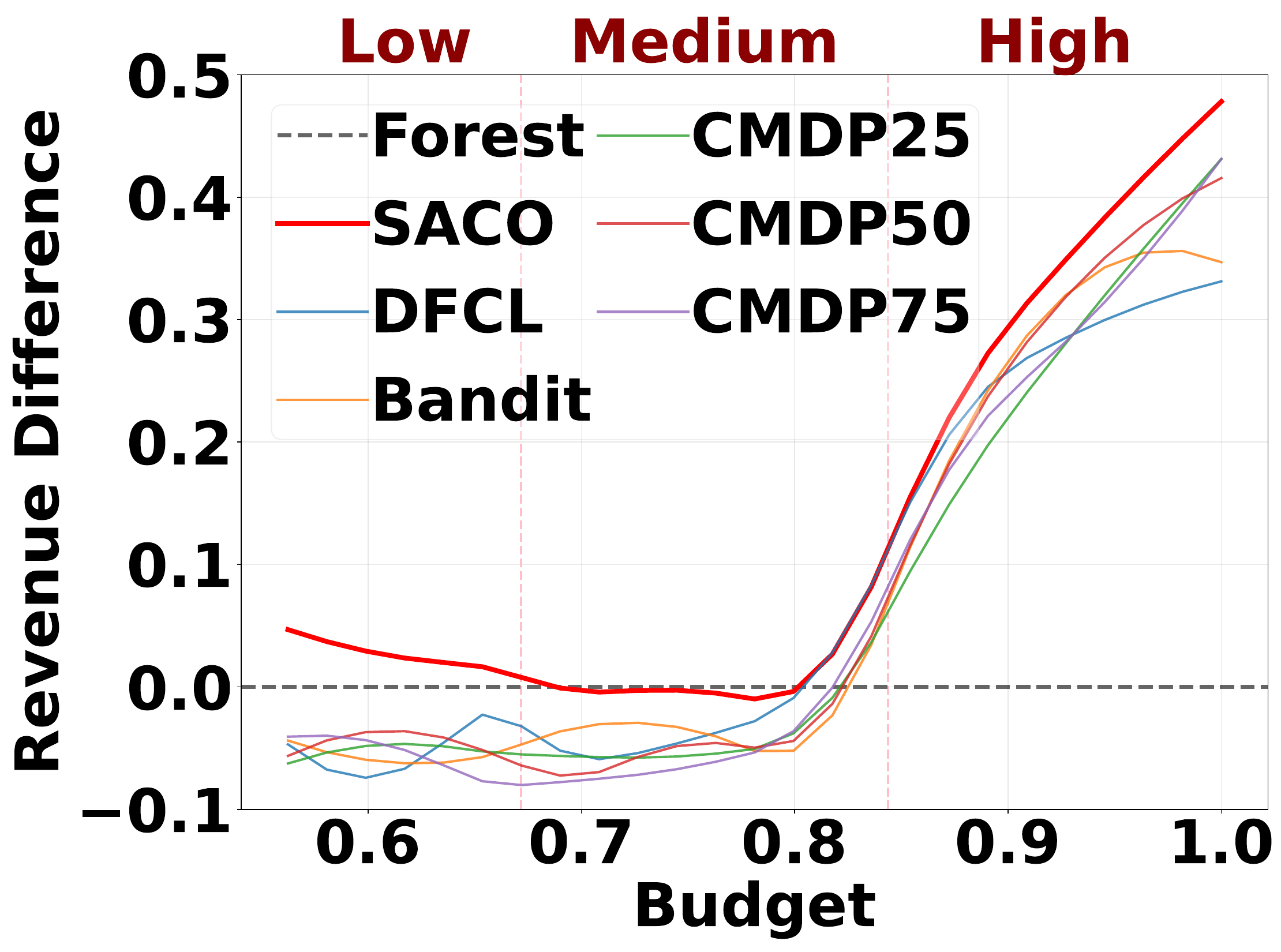}
        \caption{SY}
    \end{subfigure}
    \hfill
    \begin{subfigure}[b]{0.246\textwidth}
        \includegraphics[width=\linewidth]{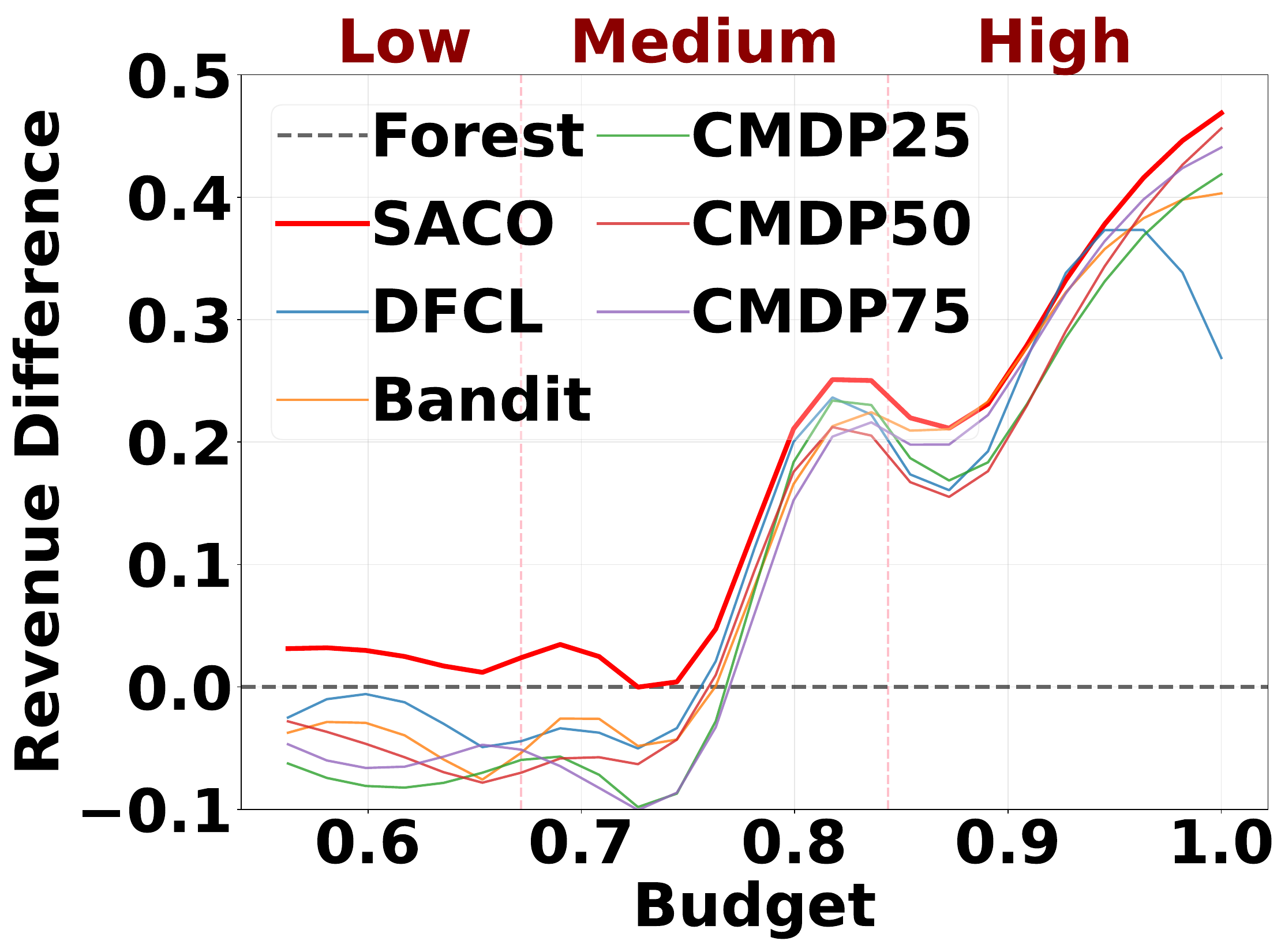}
        \caption{BD}
    \end{subfigure}

    \caption{Evaluation Results}
    \label{fig:evaluation-results}
\end{figure*}

\begin{table*}[h]
    \centering
    \scriptsize
    \label{tab:performance_comparison}
    \begin{tabular}{>{\centering\arraybackslash}m{1.2cm} >{\centering\arraybackslash}m{1.2cm} >{\centering\arraybackslash}m{1.2cm} >{\centering\arraybackslash}m{1.2cm} >{\centering\arraybackslash}m{1.2cm} >{\centering\arraybackslash}m{1.2cm} >{\centering\arraybackslash}m{1.2cm} >{\centering\arraybackslash}m{1.2cm} >{\centering\arraybackslash}m{1.2cm}}
        \toprule
        \textbf{Dataset} & \textbf{Metric} & \textbf{SACO} & \textbf{DFCL}  & \textbf{Bandit} & \textbf{CMDP25} & \textbf{CMDP50} & \textbf{CMDP75} & \textbf{Forest}\\
        \midrule
        \multirow{2}{*}{CU} & $\overline{ROI}$ &1.11&1.052&1.072&1.053&1.053&1.052&1.087 \\
                            & $\overline{BARate}$ &0.997&0.986&0.966&0.987&0.991&0.984&0.968\\
        \multirow{2}{*}{MT} & $\overline{ROI}$ &1.105&1.106&1.06&1.053&1.054&1.059&1.088 \\
                            &$\overline{BARate}$ &1.0&0.926&0.988&0.989&0.984&0.984&0.952 \\
        \multirow{2}{*}{SY} & $\overline{ROI}$ &1.109&1.073&1.066&1.053&1.053&1.057&1.101 \\
                            & $\overline{BARate}$ &0.999&0.977&0.978&0.983&0.986&0.98&0.934\\
        \multirow{2}{*}{BD} & $\overline{ROI}$ &1.126&1.069&1.067&1.05&1.057&1.052&1.101 \\
                            & $\overline{BARate}$ &0.996&0.981&0.983&0.979&0.984&0.985&0.919 \\
        \bottomrule
    \end{tabular}
    \caption{Performance Comparison Across Datasets and Baselines}
\end{table*}

In this section, we conduct experiments to evaluate the performance of our proposed framework. Specifically, we compare it with baseline algorithms, perform an ablation study to assess the contribution of each component, and conduct an inference speed analysis. Since both our framework and the baselines may generate actions $\tilde{a}$ that are not present in the dataset trajectories, the corresponding revenue $\tilde{r}$ and cost $\tilde{c}$ for these actions are missing. To address this issue of missing data, we follow the approach outlined in \cite{guo2024generative} and create an offline simulation environment based on the tuples $(s, a, r, c)$ within the existing trajectories. This environment enables us to obtain the tuples $(s, \tilde{a}, \tilde{r}, \tilde{c})$ for actions that are not included in the original trajectories.

\begin{figure*}[h]
    \centering
    \includegraphics[width=0.923\linewidth]{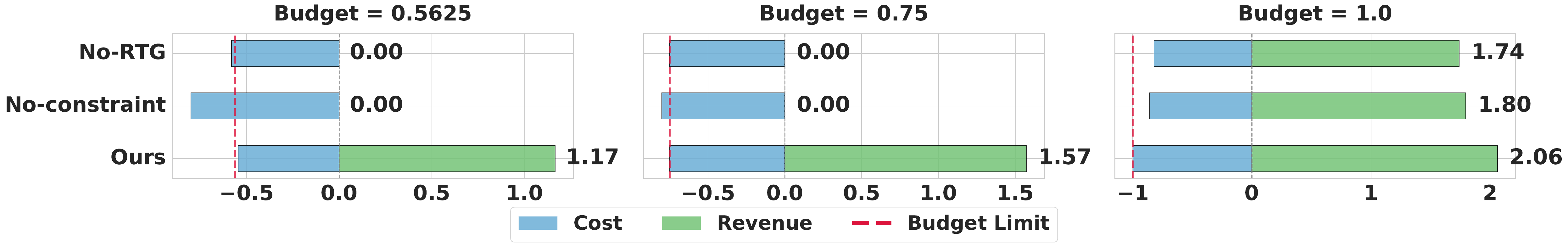}
    \caption{Ablation Study on MT Dataset.}
    \label{fig:ablation}
\end{figure*}

\subsection{Experimental Settings}
\subsubsection{Dataset}
To demonstrate the robustness of our framework, we leverage a diverse set of datasets including public benchmarks, real-world industrial data, and synthetic simulations. 

\textbf{CRITEO-UPLIFT v2 (CU).}
The dataset is provided by Criteo AI Labs at the AdKDD’18 workshop \cite{diemert2018large} and comprises 13.9 million samples collected from Randomized Controlled Trial (RCT). Each sample contains 12 features, a binary treatment indicator, and two response labels, namely visit and conversion. 
We follow the approach in \cite{zhou2023direct}, extracting user feature columns `f0-f11' as states, `treatment' as actions, `visit' label as costs and `conversion' label as generated revenue. 

\textbf{MT-LIFT (MT).}
Collected from two months of coupon marketing scenarios for food delivery\cite{huang2024entire}, the dataset contains nearly 5.5 million samples, which include 99 features, 5 types of treatment, and two response labels: click and conversion.
Similar to the CU dataset, we process user feature columns `f0-f98' as states, `treatment' as actions, `click' as costs and `conversion' as revenue. 

\textbf{Synthetic Dataset (SY).}
We have developed a synthetic dataset generation pipeline inspired by \cite{yan2023end}. The dataset consists of 10,000 simulated users. Each user interaction spans 10 timesteps, with an action space of four treatments applied randomly with equal probabilities at each timestep. The cost and reward is generated when the user responds to the coupon with the response probability. Gaussian noise is used to simulate state transitions.

\textbf{Business Dataset (BD).} To realistically evaluate the performance of our model in real-world industrial scenarios, we incorporate a recent business dataset originating from ByteDance, a global leading technology company with a range of popular platforms. User features included are anonymized to ensure data privacy and compliance. The marketing dataset consists of millions of samples, including anonymized user features as states, treatments as actions, `cost' label as costs and `is paid' label as revenue.

For all datasets, we randomly split them separately, using 80\% of the samples for training and 20\% for testing. After initial data cleaning to remove anomalies, we aggregate the data by user ID or relatively stable user feature. Within each aggregated group, we subsequently sort the data chronologically to generate trajectory samples.

\subsubsection{Evaluation Metrics}
\begin{itemize}
\item Revenue ($R(\boldsymbol{x})$): A crucial factor in measuring marketing effectiveness is the platform revenue generated by the marketing efforts. 

\item Return on Investment (ROI) ($\frac{R(\boldsymbol{x})}{C(\boldsymbol{x})}$): 
The emphasis on the ROI indicator is paramount. Without proper consideration of cost, the strategy risks favoring marketing activities targeting users with both high returns and high associated expenditures. 
It leads to a suboptimal outcome and inefficient resource allocation.

\item Budget Allocation Rate (BARate) ($\frac{C(\boldsymbol{x})}{B}$): 
Effective coupon distribution requires assessing user sensitivity beforehand. Successful strategies will exhibit high budget utilization, implying that coupons are effectively targeted at highly sensitive users. 

\end{itemize}

\subsection{Evaluation Experiments}

\subsubsection{Baseline Algorithms}
\begin{itemize}
\item Causal \textbf{Forest}\cite{athey2021policy}: This baseline model, commonly used in marketing scenarios, is trained on historical data to predict individual treatment effects (ITE). At each decision point, the model estimates the treatment effect of offering a coupon and allocates it only if the predicted effect is positive.

\item \textbf{Bandit} \cite{chapelle2011empirical}: We employ a multi-arm bandit algorithm based on Thompson Sampling (TS). This algorithm leverages Bayesian methods to construct a belief over the reward distribution of actions.

\item Constrained Markov Decision Process (CMDP) \cite{xiao2019model}: For different $\lambda$ values, the CMDP model needs to be retrained. 
Hence, we established three CMDP baselines, \textbf{CMDP25}, \textbf{CMDP50} and \textbf{CMDP75}, by retraining the core policy network with $\lambda$ values of 0.25, 0.5, and 0.75, respectively.

\item Decision Focused Causal Learning (\textbf{DFCL}) \cite{zhou2024decision}: The DCFL baseline is a state-of-the-art method that claims to achieve breakthrough performance by building upon two-stage approaches. It is an end-to-end framework, aiming to align the prediction objective with the ultimate decision-making goal.

\end{itemize}

\subsubsection{Experimental Results}
To establish a clear baseline, we select the Forest algorithm as our benchmark model. Figure \ref{fig:evaluation-results} presents the revenue difference between evaluated models and the causal forest benchmark. We choose Forest due to its widespread adoption in marketing platforms, known for its robust causal learning capabilities. It is an appropriate and meaningful point of comparison.

To assess budget constraint robustness, we defined three budget levels: Low, Medium, and High, representing strict to loose constraints. Within each level, we specify three distinct budget amounts. All budget values are normalized to avoid dependence on specific numerical scales and ensure generalizability across different scenarios.

The experimental results in Figure \ref{fig:evaluation-results} demonstrate that our proposed SACO framework gains higher revenue compared to baselines across all datasets and budget levels, with two exception: the medium budget level in the SY dataset and MT dataset. SACO achieves an average revenue improvement of \textbf{3.60\%}. Furthermore, Table 1 highlights SACO's superior performance in terms of both average ROI and BARate across the tested budget range. This overall superior performance emphasizes SACO's strong strategy learning capabilities, robust causal reasoning, and adaptability to diverse dataset characteristics.

We hypothesize that SACO's slightly lower revenue than DFCL at the medium budget level in MT is because DFCL, developed by Meituan, may be particularly suited to the Meituan dataset.
Regarding the slightly lower revenue compared to Forest at the medium budget level in the SY dataset, this is attributed to the uniform distribution of actions within that specific synthetic dataset. It is worth noting that forest and uplift models often benefit from training on unbiased RCT data, which features a uniform distribution of actions. However, online deployment would typically utilize an optimized strategy like DFCL algorithm, rather than a random coupon distribution, which means the newly sampled data inherently introduces selection bias. While some existing methods don't detail how their models would be iteratively updated under these biased conditions, our SACO's implicit causal learning approach offers a distinct advantage. By reducing its reliance on the statistical premise of unbiased data, SACO allows for efficient, timely, and cost-effective iterative updates, contributing to sustained optimal performance in dynamic online environments.

\subsection{Ablation Study}

\subsubsection{Ablation Design}
\label{Ablation-Design}
\begin{itemize}
\item No-constraint: We omit the parameter $\lambda$ and $\widehat{RTC}$ from our model,  assessing their significance on learning and respecting budget constraints.
\item No-$\widehat{RTG}$: We utilize the original Transformer model \cite{vaswani2017attention} without the $\widehat{RTG}$ parameter to evaluate its impact on the model's capacity to pursue long-term revenue, as opposed to immediate rewards.
\end{itemize}

\subsubsection{Experimental Results}
We conduct ablation studies in the MT-LIFT public dataset. The experimental results are presented in Figure \ref{fig:ablation}. The horizontal axis depicts revenue (positive values) and costs (negative values), while the vertical axis lists the ablated algorithms. The dashed vertical line denotes the budget constraint.

Comparing our framework with No-constraint model reveals that removing constraint-related variables leads to a diminished awareness of budgetary limitations. The model fails to adequately respond to budgetary changes.
Comparing SACO to model without $RTG$ shows that, when the budget is not sufficiently generous, the absence of $RTG$ causes the model to marginally exceed the budget in pursuit of immediate revenue gains. Furthermore, when budget is ample, the absence of RTG guidance restricts overall revenue generation, rendering its performance inferior to SACO.

\subsection{Inference Speed Analysis}

Motivated by the demands of real-world applications, we conduct inference efficiency experiments to validate the runtime performance of SACO. The results are presented below.

\begin{table}[h]
\centering
\label{tab:inference_speed}
\begin{tabular}{lccc}
\toprule
\textbf{Scenarios} & \textbf{Time (ms)} & \textbf{Hardware} \\
\midrule
Single-user & 24.6243  & CPU-only \\
Single-user & 3.8128  & NVIDIA RTX 4090 \\
128 users & 48.5075  & NVIDIA RTX 4090 \\
\bottomrule
\end{tabular}
\caption{Model Inference Time}
\end{table}

Based on the results, SACO framework achieves an inference speed suitable for industrial applications. 
In addition, we use 16 parallel models in the multi-user scenario, limited by CUDA memory. We anticipate that deployment in industrial environments, equipped with superior computing resources (e.g., larger GPU memory, parallel GPUs), will effectively mitigate this memory bottleneck. 

\section{Conclusion}

In this paper, we propose SACO framework, sequentially formulating coupon distribution strategies to enhance long-term revenue.
Our unified framework combines general scenarios, sequential modeling and efficient iterative updates.
Extensive experiments have been conducted to demonstrate the superiority of our framework.

Future directions includes integrating SACO with externalities, such as incorporating user-to-user impact evaluation as input features. Also, we plan to encode coupons to prevent action space drift. This is in case we need to dynamically adapt coupon types (e.g., larger discounts) based on observed coupon performance.

\section{Acknowledgments}
This work was supported by National Natural Science Foundation of China (No.62472428), Public Computing Cloud, Renmin University of China, the fund for building world-class universities (disciplines) of Renmin University of China and ByteDance.

\bibliography{aaai2026}

\end{document}